\documentclass{article}
\usepackage{ijcai17}

\usepackage{times}

\usepackage{times,helvet,courier}
\usepackage{graphicx,color,balance}
\usepackage{amssymb,amsmath,array,amsfonts,mathrsfs}
\usepackage{pgf}
\usepackage{multicol}
\usepackage{fixltx2e}
\usepackage{placeins}
\usepackage{tikz}
\usepackage{url}
\usepackage{tikzscale}
\usetikzlibrary{arrows} 
\usetikzlibrary{shapes.geometric}
\usetikzlibrary{shapes.arrows}

\usepackage[font=small,skip=0pt]{caption}

\usetikzlibrary{arrows,automata}
\usetikzlibrary{fit}					
\usetikzlibrary{backgrounds}	
\usetikzlibrary{decorations.pathmorphing,backgrounds,positioning,fit,petri}

\usepackage{ifthen}

\usepackage[textsize=scriptsize,textwidth=1cm]{todonotes}

\usepackage{braket}

\begin{document}

\title{Explainable Planning
}
\author{Maria Fox, Derek Long, Daniele Magazzeni\\
King's College London\\
\textit{firstname.lastname@kcl.ac.uk}
}
\maketitle

\begin{abstract}
As AI is increasingly being adopted into application solutions, the challenge of supporting interaction with humans is becoming more apparent. Partly this is to support integrated working styles, in which humans and intelligent systems cooperate in problem-solving, but also it is a necessary step in the process of building trust as humans migrate greater responsibility to such systems. The challenge is to find effective ways to communicate the foundations of AI-driven behaviour, when the algorithms that drive it are far from transparent to humans. In this paper we consider the opportunities that arise in AI planning, exploiting the model-based representations that form a familiar and common basis for communication with users, while acknowledging the gap between planning algorithms and human problem-solving.
\end{abstract}

\section{Introduction}\label{sec:int}


DARPA recently launched the \textit{Explainable AI (XAI) program}\footnote{http://www.darpa.mil/attachments/DARPA-BAA-16-53.pdf} that aims to create a suite of AI systems able to explain their own behaviour. This program is mainly concerned with machine/deep learning techniques, as they are currently treated almost as a black box. For example, it is not possible to fully understand why alphaGo selected a specific move at each turn, or on what basis a neural network recognises an image as an ``image of a cat". 

The need for explainable AI is motivated mainly by three reasons:
\begin{itemize}
\item the need for trust; 
\item the need for interaction; 
\item the need for transparency.
\end{itemize}

If doctors want to use a neural network to make a diagnosis, they need to be confident that there is a clear rationale for the NN to diagnose a cancer, in order to build \textit{trust}. As autonomy gathers traction, in many scenarios, instead of full autonomy,   Human-Autonomy Teaming (HAT) is required, where humans interact with the AI systems, and for this humans need to understand why the AI system is suggesting something that the human would not do: this requires \textit{interaction}. There are growing legal implications in the use of AI, and in the cases where the AI system makes the wrong decision, or simply disagrees with the human, it is important to understand why a wrong or different decision was made: this is \textit{transparency}.

Explainable AI is harder to achieve than the good decision-making that underlies it. The need to explain decisions forces them to be made in ways that can be subsequently justified in human terms. Entirely trustworthy and theoretically well-understood algorithms can still yield decisions that are hard to explain. For example, linear programming is a well-established tool, but explaining the results it generates without simply `appealing to authority' remains hard. Part of the difficulty lies in understanding what an explanation should actually contain. 


On one hand it is evident that there has been amazing progress in machine/deep learning research and there is a huge proliferation of ML and DL learning. On the other hand, Deep Neural Networks are still far away from being explainable.

In contrast, AI Planning is potentially well placed to be able to address the challenges that motived the DARPA project on AI: planners can eventually be trusted; planners can allow an easy interaction with humans; planners are transparent (at least, the process by which the decisions are made are understood by their programmers). 

This paper presents Explainable Planning (XAIP), describing some initial results, and proposing a roadmap for making XAIP more effective and efficient.

Of course the challenge of Explainable AI and the  need of making machine/deep learning explicable remain of critical importance. At the same time, we think that XAIP is an important contribution in this direction, as Planning is an important area of AI with applications in domains where learning is not an option.

The paper is structured as follows. We give an overview of related work in the next section. In Section 3 we  list  some important questions that XAIP should address and in Section 4 we discuss the features of planning that facilitate explanations. In Section 5 we present initial results and suggest future directions. In Section 6 we show two illustrative examples.
Section 7 concludes the paper.




\section{Related Work}

For a survey of recent works in the broader area of Explainable AI, we refer to the IJCAI-17 XAI workshop website\footnote{http://home.earthlink.net/$\sim$dwaha/research/meetings/ijcai17-xai/}.
Here we briefly highlight some recent works that are related and can contribute to Explainable Planning. \textit{Plan Explanation} is an area of Planning where the main goal is to help humans to understand the plans produced by the planners (e.g., \cite{Sohrabi}. This involves the translation of the planner outputs (e.g., PDDL plans) in forms that humans can easily understand; the design of interfaces that help this understanding (e.g., spoken language dialog systems~\cite{biundospoken}); and the description of 
causal and temporal relations for plan steps (e.g., \cite{biundo}). Note that making sense of a plan (plan explanation) is different from explaining why a planner made decisions (XAIP).

\textit{Plan Explicability}~\cite{rao2} focuses on human's interpretation of plans. Learning is used to create a model of the  interpretations, which is then used to measure the explicability and predictability of plans.

Veloso and her team look at the problem of generating narrations for autonomous mobile robot navigations. They contribute with \textit{verbalization}, where the robot experience is described via natural language~\cite{veloso1}.

In \textit{Model reconciliation}~\cite{rao1}, the focus is on the agent and the human having two different models, hence the explanations must identify and reconcile the relevant differences between the models.

David Smith in his AAAI invited talk presented \textit{Planning as an Iterative Process}~\cite{smith12}, and he discussed the broad problem of users interacting with the planning process, which also includes questions about choices made by the planner. Pat Langley et al. more recently used \textit{Explainable Agency} to refer to the ability of autonomous agents to explain their decisions, and in  ~\cite{langley17} they discuss some functions that agents should exhibit.  

In this paper, we go beyond a discussion of the questions that need to be answered, and by focusing on AI planning we provide initial results on how to address some of the questions and we point to concrete works in the community to address the others.

\section{Things to Be Explained}\label{sec:questions}

As mentioned in the introduction, one of the challenges of XAI is to understand what constitutes an explanation. In general, rewriting the steps of the decision-making algorithm in natural language is not what is required. For example, despite it being the case that many planners select actions in their plan-construction process in order to minimize a heuristic distance to goal, based on a relaxed plan, even these terms are inappropriate vocabulary to explain the process to a human. In any case, a real danger in XAI is to reduce an explanation to the statement of the obvious. It is clearly not the answer to the question `why did you do that?' to say 'because it got me closer to the goal'. A request for an explanation is really an attempt to uncover a piece of knowledge that the questioner believes must be available to the system and that the questioner does not have.

In this section we list some of the questions that characterise what it means for the behaviour of a planner to be \textit{explainable}, and we discuss what constitutes a response to these questions. 

\begin{itemize}

\item Q1: Why did you do that?

This is one of the most fundamental questions that can be asked about a plan. It is also an excellent example of how complex the intention behind the question can be. In a sufficiently long and complex plan, it is plausible that the questioner is unable to immediately see which later action in the plan is supported by the target action. Thus, the answer could be as simple as `action A is in the plan to allow this application of action B'. However, for shorter and more easily assimilated plans, the question is far more likely to be an implicit question: `why did you do action A? {\em I would have done action B}' 

\item Q2: And why didn't you do \textit{something else} (that I would have done)?

This question is similar to the intention in Q1, but makes the alternative action explicit. An answer to this question would normally be a demonstration of a flaw in a plan that uses the proposed alternative action compared with the plan actually produced. It would usually be acceptable to demonstrate that the plan actually produced was no worse than a plan using the proposed alternative action. In order to respond with either a flaw or else a demonstration of neutral cost, it is necessary to infer by what metric the alternatives are to be compared (one plan might be longer but cheaper than a second --- depending on the relative values of time and money, either plan might be considered better).

\item Q3: Why is what you propose to do more efficient/safe/cheap than something else (that I would have done)?

This question refines Q2 by being explicit about the metric being used to evaluate the plans. If the metric is different to the one used in constructing the original plan, then the answer might be to point out the different basis for evaluation of plans. This is a valid explanation provided the original plan is better under the original metric than the rival proposal.  

\item Q4: Why can't you do that?

Here we consider the form of this question arising when a planner fails to find a plan for a problem. Planners are typically not very effective at proving unsolvability of planning problems, but model-checking techniques can be applied to planning domain models in order to attempt to prove the non-existence of plans. Unfortunately, converting the exhaustive search of a space (albeit supplemented with careful relaxation-based approaches that bundle parts of the search space) into a transparent and succinct argument is extremely challenging. 

\item Q5: Why do I need to replan at this point?

During execution, plan failure will be caused by a deviation between the expected behaviour and the observed behaviour of the world. This question can be directed at discovering what has diverged from expectation, or it might be that the deviation is understood, but the significance of the deviation is not. Thus, this question seeks to know what is it that the executing plan was depending on being true that has been observed not to be.

\item Q6: Why do I not need to replan at this point?

There are two reasons possible for this question. One is that the observer has seen a divergence in expected behaviour and does not understand why it should not cause plan failure and the other is that the observer has seen non-diverging behaviour that is not what was expected by the observer. In other words, the divergence could be observable for the executive (and then the explanation is to show why it does not cause plan failure) or else there might be no divergence observable by the executive, in which case the explanation (assuming there is a valid explanation) is to show why the observed behaviour was expected at this point.



\end{itemize}

Of course there are other questions related to Explainable Planning, including those arising when we consider probabilistic planning or planning under uncertainty, as well as anytime planning (e.g., \textit{will I get a significantly better plan if I give the planner 10 more minutes?}).

\section{Unique Features of Planning}


AI Planning exploits a collection of techniques that have the potential to make it easier to understand the decision process (even if it is very complex and requires sophisticated algorithms and heuristics).

First of all, AI Planning is based on \textit{models}. Models are used to create plans, and can also be used during plan execution as well as after a plan has been executed. One of the driving principles for a large part of the work carried out in planning is that planners should encapsulate the machinery of planning independently of the domain of application. This means that models capture the dynamics of domains. However, a second guiding principle has been that domain models should not attempt to direct the decision process in the planner (this is not a universally adopted principle, but it has been a consequence of the international planning competition series that domain models have been developed to be `pure' descriptions of what can be done, not how to do it). McDermott articulated this as the maxim that domain models should contain `axioms, not advice'. The implication of this for modellers is that they do not need to understand how a planner works, but only the behaviour of their domain. This leads to more intuitive and accessible models for users and facilitates their use in explanation.

Second, plan execution provides its \textit{execution trace}, as a set of pairs (observation, action) which can be used to explore the reasons behind the choices of actions and allows explanations to focus on aspects of state or of action choice, depending on the question.

Third, some progress has been made in explaining plans, in order to help humans to understand the meaning of plans, with a long history based on mixed-initiative planning.

Most planners are based on transparent algorithms, where the planner choice at each decision point is deterministic, repeatable and based on a specific choice mechanism. The fact that the reason for the choice of an action is transparent to the programmer, at least, makes it plausible that we can construct an articulation of parts of that reason in a form a human user might appreciate.


\section{Providing Explanations}
In this section we address the questions presented in Section~\ref{sec:questions}. For each question, we highlight the main challenges, present some preliminary results and propose a roadmap for achieving the goal of providing reasonable answers and explanations.

\subsection{Explaining why the planner chose an action}
This explanation introduces two main challenges, that are also inherited by all the following explanations. First, an explanation needs to show \textit{causality} among actions. While this is obvious in many cases (e.g., I get the key first so that I can open the door later), there are examples where action A early in the plan is needed to support action B much later in the plan. This causal relationship might not be evident.

As a practical example, in the electricity domain~\cite{chiara}, the planner reverses current through a transformer \textit{early} in the afternoon in order to support the achievement of supply within thermal constraints in the peak demand period \textit{later} in the evening. This was performed early because there was more flexibility to reconfigure the flows when the load was lighter and by the time the supply was required, demand was so high that the network had no longer got flexibility to change the configuration.

The second issue is that the plan must be understandable to humans, who are not supposed to be planning researchers or experts. Hence, planning formalisms such as PDDL need to be presented to humans in a more natural-language fashion, and the works on plan explanation go in this direction (e.g., \cite{biundo}).


\subsection{Explaining why the planner did not choose an action}

When a planner decision is confronted with an alternative suggested by the human, an explanation should be a demonstration that the alternative action would prevent from finding a valid plan or would lead to a plan that is no better than the one found by the planner. 
However, just showing the different heuristic value is not a valid explanation, unless it is translated into a value which is more informative for the user than for example the RPG heuristic value. 

What is needed, instead, is an algorithm that executes the plan up to the point where the human suggests the alternative, then injects the human decision, and finally replans from the state obtained after applying the action suggested by the human.

\begin{figure*}[ht]
  \centering
\begin{tikzpicture}[>=latex]
   
  \begin{scope}


  \filldraw[black] (0,0) circle (2pt) node[anchor=south] {s};
  \node (p0_d) at (-0.4,-0.4) {A};
  
  \filldraw[black] (-0.5,-1.0) circle (2pt) node[anchor=east] {$s_1$};
  
  \draw[<-] (-0.4,-0.9) -- (0.0,0.0);
  \draw[<-] (0.4,-0.9) -- (0.0,0.0);

  \filldraw[black] (0.5,-1.0) circle (2pt) node[anchor=west] {$r_1$};
\node (p0_d) at (0.4,-0.4) {B};
\draw [->,dashed,blue] (0.5,-1.0) to [bend left=45] (0.0,-0.2);
  
  \filldraw[black] (-0.5,-4.3) circle (2pt) node[anchor=north] {$g_{A}$};

\draw [->,decorate,decoration=snake] (-0.5,-1.0) -- (-0.5,-4.3);

  \filldraw[black] (3.5,0) circle (2pt) node[anchor=south] {s};
  \node (p0_d) at (3.1,-0.4) {A};
  
  \filldraw[black] (3,-1.0) circle (2pt) node[anchor=east] {$s_1$};
  
  \draw[<-] (3.1,-0.9) -- (3.5,0.0);
  \draw[<-] (3.9,-0.9) -- (3.5,0.0);

  \filldraw[black] (4.0,-1.0) circle (2pt) node[anchor=west] {$r_1$};
\node (p0_d) at (3.9,-0.4) {B};

  \filldraw[black] (2.75,-2.0) circle (2pt) node[anchor=east] {$s_2$};
  \draw[<-] (2.75,-1.9) -- (3.0,-1.0);
  \node (p0_d) at (2.65,-1.5) {$\alpha_1$};

  \filldraw[black] (4.25,-2.0) circle (2pt) node[anchor=west] {$r_2$};
  \draw[<-] (4.25,-1.9) -- (4.0,-1.0);
  \node (p0_d) at (4.35,-1.5) {$\beta_1$};
  
  \node (dotsl) at (2.55,-2.72) {$\cdots$};
  \node (dotsl) at (4.4,-2.72) {$\cdots$};

    \draw[<-] (2.6,-2.5) -- (2.75,-2.0);
    \draw[<-] (4.35,-2.5) -- (4.25,-2.0);

  
    \draw[<-] (2.4,-3.3) -- (2.55,-2.8);
    \node (p0_d) at (2.3,-2.9) {$\alpha_k$};

    \draw[<-] (4.6,-3.3) -- (4.45,-2.8);
    \node (p0_d) at (4.75,-2.9) {$\beta_k$};
    \filldraw[black] (4.6,-3.4) circle (2pt) node[anchor=west] {$r_{k+1}$};

  \filldraw[black] (2.4,-3.4) circle (2pt) node[anchor=west] {$s_{k+1}$};
\draw [->,decorate,decoration=snake] (2.4,-3.4) -- (2.4,-4.2);
  \filldraw[black] (2.4,-4.3) circle (2pt) node[anchor=north] {$g_{A}$};

   \draw[->,dashed,blue] (4.6,-3.4) to [bend left=30]  (2.45,-3.4);
   \draw[->,dashed,blue] (4.6,-3.4) to [bend right=30]  (2.8,-2.0);
   \draw[->,dashed,blue] (4.6,-3.4) to [bend right=0]  (2.7,-2.72);

  \filldraw[black] (7.0,0) circle (2pt) node[anchor=south] {s};

\node (p0_d) at (6.6,-0.4) {A};
  
  \filldraw[black] (6.5,-1.0) circle (2pt) node[anchor=east] {$s_1$};
  
  \draw[<-] (6.6,-0.9) -- (7.0,0.0);
  \draw[<-] (7.4,-0.9) -- (7.0,0.0);

  \filldraw[black] (7.5,-1.0) circle (2pt) node[anchor=west] {$r_1$};
\node (p0_d) at (7.4,-0.4) {B};

  \filldraw[black] (6.5,-4.3) circle (2pt) node[anchor=north] {$g_{A}$};

  \filldraw[black] (7.5,-4.3) circle (2pt) node[anchor=north] {$g_{B}$};
  
  \draw [->,decorate,decoration=snake] (6.5,-1.0) -- (6.5,-4.3);
  \draw [->,decorate,decoration=snake] (7.5,-1.0) -- (7.5,-4.3);
  

  \filldraw[black] (10.0,0) circle (2pt) node[anchor=south] {s};

\node (p0_d) at (9.6,-0.4) {A};
  
  \filldraw[black] (9.5,-1.0) circle (2pt) node[anchor=east] {$s_1$};
  
  \draw[<-] (9.6,-0.9) -- (10.0,0.0);
  \draw[<-] (10.4,-0.9) -- (10.0,0.0);

  \filldraw[black] (10.5,-1.0) circle (2pt) node[anchor=west] {$r_1$};
\node (p0_d) at (10.4,-0.4) {B};

  \filldraw[black] (9.5,-4.3) circle (2pt) node[anchor=north] {$g_{A}$};

  
  \draw [->,decorate,decoration=snake] (9.5,-1.0) -- (9.5,-4.3);
  \draw [->,decorate,decoration=snake] (10.5,-1.0) -- (11.0,-2.25);

\draw (11,-2.25) -- (12,-3.6) -- (10,-3.6) 
  -- cycle;

\node (a) at (0,-4.8) {(a)};
\node (b) at (3.5,-4.8) {(b)};
\node (c) at (7.0,-4.8) {(c)};
\node (d) at (10.0,-4.8) {(d)};

  
  
     

      
     
%

%
%
%
  \end{scope}
\end{tikzpicture}
  \caption{Possible plan behaviours after human-decision injection.}
  \label{fig:planbehaviour}
\end{figure*}
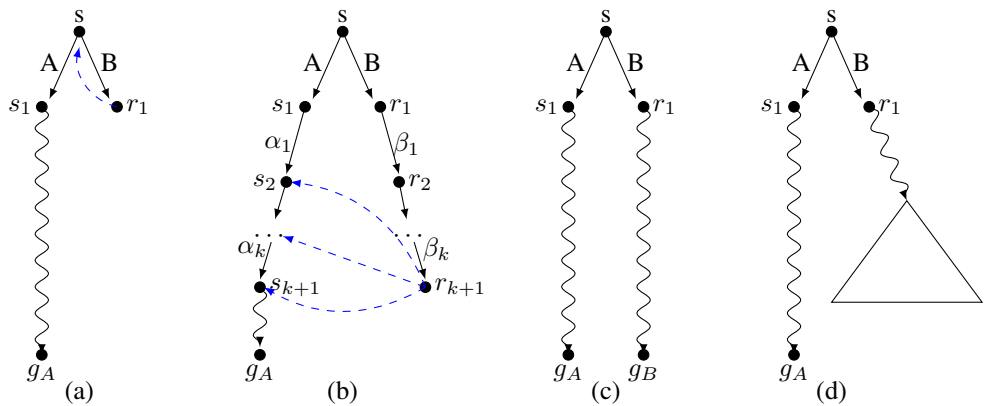

Here there is a first issue, though, as one possible behaviour of the planner could be to just do an \textit{undo} of the human action and produce the original plan (see Figure~\ref{fig:planbehaviour} part (a)).

One possible fix is to forbid the planner to revisit the state where the human decision was injected. This, however, does not prevent the planner to get back to the original plan after $k$ steps, as shown in Figure~\ref{fig:planbehaviour} part (b). In this case, the explanation can provide the different costs of the two alternatives, that is $C_A= c (A)+c(\alpha_1)+\ldots+c(\alpha_k)$ and $C_B= c(B)+c(\beta_1)+\ldots+c(\beta_k)$.
More in general, the human may want to inject a longer plan, and in this sense the explanation must enable the interaction with the human, by allowing the human to inject more than one action, and provide an explanation after each injection.

The two remaining possible outcomes after the human-decision injections are either that the planner finds a plan for a different goal, or it fails to find a plan, as shown in Figure~\ref{fig:planbehaviour} parts (c) and (d), respectively.

\subsection{Explaining why the planner decisions are \textit{better}}
The third question we listed was: \textit{why is the planner decision more efficient/safe/cheap than what I would do?} The focus here is that different \textit{metrics} can be used to evaluate a plan. The more interesting case is when one wants to evaluate a plan using a metric which is different from the one used when searching for the plan. This is of practical importance, given that when dealing with complex domains (temporal/numeric domains) there are a number of planners able to minimise the makespan, while almost none are able to optimise other metrics (e.g., total cost). Hence, for example one would like to understand whether a plan found using POPF or FastDownward is actually more efficient (in terms of total cost), than an alternative plan suggested by the human.

This explanation is a refinement of previous explanations, and our proposed solution is to integrate the previous explanation approach with the validator VAL~\cite{val}, where different metrics can be specified to evaluate plans. After each injection of human decision, the validator is used to execute the alternative plan against the new metric (total cost for example).

\subsection{Explaining why things can \textit{not} be done}
There are two reasons why one action can not be applied in a given state. Either because the current state does not satisfy the action precondition; or because the application of that action would prevent achieving the goal from the resulting state.
To provide an explanation for the first case is straightforward, and the validator VAL already provides this.
Explanations for the second case are more challenging, and anyway would also be used for providing more general justifications for why the goal cannot be achieved at all (i.e., the planning problem is unsolvable). To this end, we suggest that a promising direction is given by the work being done in proving plan non-existence (see, among others, \cite{Backstrom}, \cite{hoffmann1}, \cite{hoffmann2}).

Model-checking algorithms and tools can prove very suitable~\cite{mc}. Indeed, in the planning-as-model-checking paradigm~\cite{aipasmc}, a planning problem is cast as a verification problem, where the safety property to be verified is set as the negation of the goal of the planning problem. In this way, if the model checker returns an error trace, that would correspond to the plan. On the contrary, if the model checker states that the property (not goal) is satisfied for all the reachable states, this is a proof that there is no plan. Recently, this paradigm has been applied to complex planning problems with temporal and numeric features~\cite{bogom,bog+}.

Another very relevant topic is the research around Simple Temporal Networks~\cite{stn}. STN can be used, for example, to show why an action taking longer than expected can invalidate the plan. There is a large amount of research around STN, STNU~\cite{stnu}, and controllability of STN. For a survey on this field, we refer to \cite{micheli}.

While proving plan-non-existence (or STN inconsistency) is not yet explaining why the problem is unsolvable, we highlight here that a roadmap for this question should build on the cited works.

\subsection{Explaining why one needs to replan}\label{sec:filter}
This kind of explanation is needed at plan-execution time. In many real-world scenarios, it is not obvious that the plan being executed will fail for some changes in the environment and/or for mismatching about the model of the environment and the real environment. In most of the cases, plan failure is discovered only when it is too late for replanning in an efficient way. 

Explanations for when replanning is needed must take into account the whole plan being executed and check its validity when monitoring the environment. 
To this end, one possible approach is to use the \textit{filter violation} techniques, as described in~\cite{cas15} and implemented in ROSPlan.

ROSPlan uses a Knowledge Base to store information about the environment and the plan being executed. The Knowledge Base is updated as soon as new information
becomes available. Each change to the Knowledge Base is checked against a filter that is created as follows.
Once the plan is generated, the filter is constructed by taking the intersection
of static facts in the problem instance with the union
of all preconditions of actions in the plan. In addition, each
object instance involved in these facts is added to the filter.
For example, suppose we have a PDDL domain with the object type \texttt{waypoint}, the static fact \texttt{(connected ?from ?to - waypoint)} and the action \texttt{navigate ( ?v − vehicle ?from ?to − waypoint)}  whose preconditions include \texttt{(connected ?from ?to)}. For each \texttt{navigate} action scheduled in the
plan, the waypoint instances bound to \texttt{from} and \texttt{to} and the ground fact \texttt{(connected ?from ?to)} are included in the filter. If these objects are removed, or altered in the Knowledge Base, a notification will be sent to the Planning System.
An example of this type of explanation is provided in Section~\ref{example}.

Another promising approach in this direction is DiscoverHistory, described in~\cite{molineauxKK12}. 

\subsection{Explaining why one does \textit{not} have to replan}
This explanation is of practical importance, as it is concerned with the situation where the environment being observed (including the plan execution) is different from what was anticipated. In this very common case, it is important to avoid the naive approach of continuous replanning. Rather, it would be ideal to have a way to understand why the plan is still valid, despite the differences between what expected and what being observed. 
The most common scenario is when actions are taking longer than expected (again, one can think of an underwater mission, where an unexpected current is slowing down the AUV, hence navigate actions are taking longer than anticipated). We highlight here that a promising research direction is represented by the work on dynamic controllability of STN (e.g., \cite{dyncon}, \cite{muscettola}).

\section{Illustrative Examples}\label{example}
In this section we provide two examples of how the approach described in the previous section can be used to explain a plan and to explain why replanning is needed.

\subsection{The Rover Domain}

We consider the \texttt{rover time} domain from IPC-4 and problem 3. Here is the plan found by POP-F~\cite{popf}.

\begin{scriptsize}
\begin{verbatim}
0.000: (navigate r1 wp3 wp0)  [5.0]
0.000: (navigate r0 wp1 wp0)  [5.0]
5.001: (calibrate r1 camera1 obj0 wp0)  [5.0]
5.001: (sample_rock r0 r0store wp0)  [8.0]
10.002: (take_image r1 wp0 obj0 camera1 col)  [7.0]
13.001: (navigate r0 wp0 wp1)  [5.0]
17.002: (navigate r1 wp0 wp3)  [5.0]
18.001: (comm_rock_data r0 general wp0 wp1 wp0)  [10.0]
22.003: (navigate r1 wp3 wp2)  [5.0]
27.003: (sample_soil r1 r1store wp2)  [10.0]
28.002: (comm_image_data r1 general obj0 col wp2 wp0) [15.0]
43.003: (comm_soil_data r1 general wp2 wp2 wp0)  [10.0]

[Duration = 53.003]
\end{verbatim}
\end{scriptsize}
An instance of Q1 could be: \textit{"Why did you use Rover0 to take the rock sample at waypoint0?"}

A naive answer could be: \textit{so that I can} \texttt{communicate\_rock\_data} \textit{from Rover0 later in the plan (at 18.001)}. This is naive because the plan is so short that the user can easily see that this is performed and, presumably, will realise that the data can only be communicated by the rover that has it. A better way to interpret the question would be to consider alternative ways to achieve the goal this action supports: to communicate the rock data from Waypoint0. It turns out the only way to communicate the rock data is to first have the rock analysis from Waypoint0. And there are only two ways to do this, either to sample rock with Rover0 and or with Rover1. 

Hence, an instance of Q2 could be: \textit{Why didn't Rover1 take the rock sample at waypoint0?}
In order to provide an answer to this question, we need to force the planner to second the human input. To this end, we remove the ground action instance for Rover0 from those available to the planner and ask it to replan, and here is the new plan:

\begin{scriptsize}
\begin{verbatim}
0.000: (navigate r1 wp3 wp0)  [5.0]
5.001: (calibrate r1 camera1 obj0 wp0)  [5.0]
10.002: (take_image r1 wp0 obj0 camera1 col)  [7.0]
10.003: (sample_rock r1 r1store wp0)  [8.0]
18.003: (navigate r1 wp0 wp3)  [5.0]
18.004: (drop r1 r1store)  [1.0]
23.004: (navigate r1 wp3 wp2)  [5.0]
28.004: (comm_image_data r1 general obj0 col wp2 wp0) [15.0]
28.005: (sample_soil r1 r1store wp2)  [10.0]
43.005: (comm_soil_data r1 general wp2 wp2 wp0)  [10.0]
53.006: (comm_rock_data r1 general wp0 wp2 wp0)  [10.0]

[Duration = 63.006]
\end{verbatim}
\end{scriptsize}

Clearly this is far worse quality than the first plan (the metric is specified as makespan for these plans). So the answer could be: \textit{Because not using Rover0 for this action leads to a worse plan}. It could be argued that this is not a very satisfactory answer, although it is better than the naive answer above, because it does not seem to explain why Rover1 does everything.

One option is for the human to follow up with another question:
\textit{Why does Rover1 do everything?}
In order to answer this question, we can require the plan to contain at least one action that has Rover0 as an argument. This could be encoded in the domain automatically, by adding a dummy effect to all actions using Rover0 and then adding this as a goal, but here we use a plan generated by remodelling the domain manually, and this is the new plan found by the planner:

\begin{scriptsize}
\begin{verbatim}
0.000: (navigate r0 wp1 wp0)  [5.0]
0.000: (navigate r1 wp3 wp0)  [5.0]
5.001: (calibrate r1 camera1 obj0 wp0)  [5.0]
10.002: (take_image r1 wp0 obj0 camera1 col)  [7.0]
10.003: (sample_rock r1 r1store wp0)  [8.0]
18.003: (navigate r1 wp0 wp3)  [5.0]
18.004: (drop r1 r1store)  [1.0]
23.004: (navigate r1 wp3 wp2)  [5.0]
28.004: (comm_image_data r1 general obj0 col wp2 wp0)  [15.0]
28.005: (sample_soil r1 r1store wp2)  [10.0]
43.005: (comm_soil_data r1 general wp2 wp2 wp0)  [10.0]
53.006: (comm_rock_data r1 general wp0 wp2 wp0)  [10.0]
\end{verbatim}
\end{scriptsize}

Hence, an explanation is that this plan, while not being any longer, contains more actions, so is even worse than the last plan (and in fact contains all the actions of the last plan, so is actually a simple extension of that plan).

However, this is also not entirely satisfactory, because it only shows us that the planner cannot find a useful way to incorporate Rover0 into the plan, but not why. If we restrict the actions that can be used to achieve the dummy condition (that Rover0 acted in the plan) to the set of actions that achieve goals, then the planner cannot find a plan. So, the answer to the question could be slightly improved to: \textit{I cannot find a plan in which Rover0 does not sample the rock at Waypoint0, but achieves a goal in the plan}. 

In fact, it turns out that the problem specification prevents Rover0 from reaching Waypoint2, so the soil data there cannot be collected by Rover0, while only Camera1 can be calibrated for Objective0, so only Rover1 that carries Camera1 can be used to take that image. Therefore, the only task that Rover0 can perform is the rock sample mission at Waypoint0. 

\subsection{The AUV Domain}
We consider the AUV domain from the PANDORA project~\cite{icra,auro}, where an AUV has to complete an inspection mission, by navigating (\texttt{do\_hover}) between waypoints and making observation of a set of inspection points. Here is a fragment of a plan for this scenario:

\begin{scriptsize}
\begin{verbatim}
0.000: (observe auv wp1 ip3)  [10.000]
10.001: (correct_position auv wp1)  [10.000]
20.002: (do_hover auv wp1 wp2)  [71.696]
91.699: (observe auv wp2 ip4)  [10.000]
101.700: (correct_position auv wp2)  [10.000]
111.701: (do_hover auv wp2 wp23)  [16.710]
128.412: (observe auv wp23 ip5)  [10.000]
138.413: (correct_position auv wp23)  [10.000]
148.414: (observe auv wp23 ip1)  [10.000]
158.415: (correct_position auv wp23)  [10.000]
168.416: (do_hover auv wp23 wp22)  [16.710]
185.127: (do_hover auv wp22 wp26)  [30.201]
215.329: (observe auv wp26 ip7)  [10.000]
225.330: (correct_position auv wp26)  [10.000]
235.331: (do_hover auv wp26 wp21)  [23.177]
258.509: (observe auv wp21 ip2)  [10.000]
268.510: (correct_position auv wp21)  [10.000]
278.511: (do_hover auv wp21 wp27)  [21.255]
299.767: (observe auv wp27 ip8)  [10.000]
309.768: (correct_position auv wp27)  [10.000]
319.769: (observe auv wp27 ip6)  [10.000]
329.770: (correct_position auv wp27)  [10.000]
339.771: (do_hover auv wp27 wp17)  [23.597]
363.369: (do_hover auv wp17 wp25)  [21.413]
384.783: (do_hover auv wp25 wp32)  [16.710]
401.494: (do_hover auv wp32 wp36)  [21.451]
422.946: (observe auv wp36 ip9)  [10.000]
432.947: (correct_position auv wp36)  [10.000]
442.948: (observe auv wp36 ip15)  [10.000]
\end{verbatim}
\end{scriptsize}

In the PANDORA project, the plans are generated and dispatched through the ROSPlan framework~\cite{cas15} which is also used to monitor plan execution. ROSPlan implements the filter violation described in Section~\ref{sec:filter}. 

\begin{figure}[ht]
\begin{center}
\includegraphics[scale=0.47]{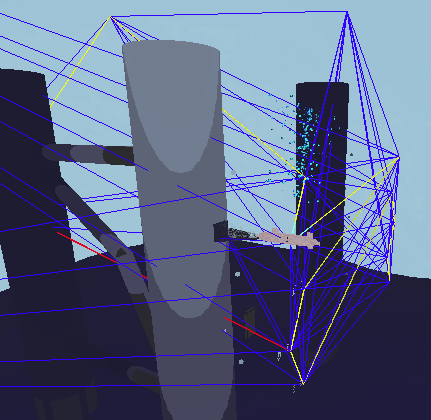}
\end{center}
\caption{Plan Execution in the AUV Domain.}
\label{fig:auv}
\end{figure}

Figure~\ref{fig:auv} shows the execution of the plan above.

The blue lines represent the Probabilistic Road Map used to determine accessibility for the AUV, while the yellow lines represent the AUV trajectory. At time-point 215.329, while the AUV is observing the inspection point \texttt{ip7} it also observes that waypoints \texttt{wp32} and \texttt{wp36} are actually not connected (this is represented by the red line in Figure~\ref{sec:filter}). Given that ~\texttt{(connected wp32 wp36)} was in the filter and after this observation the predicate is removed, a filter violation is triggered, which provides a justification for why replanning is needed, explaining that an action later in the plan (precisely at time-point 401.494) will no longer be executable, and also highlighting which condition has changed from its expected value to one that prevents the plan from being executable.

\section{Conclusion}\label{sec:con}
We have introduced Explainable Planning (XAIP), as a promising contribution to the Explainable AI (XAI) challenge. We characterised  some of the questions that need to be explained, and provided initial results and a roadmap for achieving the objective of providing effective explanations.
The next steps include a full formalisation of the XAIP problem, and a formulation of the user/planner interaction  in terms of new constraints to add and alternatives to explore.

This work opens up a number of future directions in explanations for plans as well as plan execution. For example temporal planning introduces interesting planning choices about the order in which (sub)goals are achieved. Another interesting problem is to understand whether to expect improvement in giving the planner a given additional amount time for planning. For plan execution, especially with probabilistic planning and planning under uncertainty, one of the problems is to explain what has been observed at execution time that made the planner make a particular choice.

There is no clear way to define what constitutes a good explanation. As we argued in the paper, XAIP should not focus on explaining the obvious. However, defining a good metric for explanation is an important issue.

More in general, the literature in planning contains many works that could contribute to Explainable Planning. On the other hand, nowadays plans are much more complex than before, and are also used in many new critical domains. Existing works should be revisited and leveraged in order to make XAIP more effective and efficient.





\clearpage
\bibliographystyle{named}
\balance
\bibliography{bibliography}

\end{document}